\documentclass[11pt]{article}

\usepackage[]{acl}

\usepackage{times}
\usepackage{latexsym}

\usepackage[T1]{fontenc}

\usepackage[utf8]{inputenc}

\usepackage{microtype}

\usepackage{inconsolata}
\usepackage{xspace}

\usepackage{graphicx}
\usepackage{booktabs}
\usepackage{multirow}
\usepackage{subcaption}
\usepackage{listings}
\usepackage{authblk}

\def\secref#1{\S\ref{sec:#1}}
\def\seclabel#1{\label{sec:#1}}

\title{Evaluating Contextually Mediated Factual Recall \\ in Multilingual Large Language Models}

\author[]{\bf{Yihong Liu}$^{1,2}$}
\author[]{\bf{Bingyu Xiong}$^{1}$}
\author[]{{\bf Hinrich Sch\"utze}$^{1,2}$}

\affil[]{$^1$Center for Information and Language Processing, LMU Munich \\$^2$Munich Center for Machine Learning (MCML)
 \protect\\ \texttt{yihong@cis.lmu.de}} 

\begin{document}
\maketitle
\begin{abstract}

Large language models (LLMs) can recall a wide range of factual knowledge across languages.
However, existing factual recall evaluations primarily assess fact retrieval in isolation, where the queried entity is explicitly named and the fact is requested directly.
In natural language use, facts are often accessed through context, where the relevant entity is introduced only \emph{indirectly}.
In this work, we study \emph{contextually mediated factual recall}, asking whether LLMs can reliably retrieve factual knowledge when the target entity is embedded in a naturalistic context rather than queried explicitly, across languages.
We construct controlled prompts that preserve the underlying fact while introducing referential mediation through contextual sentences.
To disentangle contextual effects from name-specific associations, we further compare performance using synthetic names and real names across languages.
Evaluating multiple model families in five languages, we find that contextual mediation consistently degrades factual recall, with substantial variation across relations.
Larger models are more robust to contextual mediation, exhibiting a reduced performance gap relative to direct queries, while the effect of real names and name origin is mixed and unsystematic.
These findings highlight a gap between isolated factual recall and context-dependent language understanding in multilingual LLMs.

\end{abstract}

\section{Introduction}

LLMs have demonstrated a strong ability to store and recall factual knowledge spanning various domains gained from their pretraining stage \citep{petroni-etal-2019-language,jiang-etal-2020-know}.
This capability extends beyond English, with multilingual models showing factual recall across languages \citep{jiang-etal-2020-x,kassner-etal-2021-multilingual,yin-etal-2022-geomlama}.

However, most existing evaluations assess factual recall in a highly \emph{decontextualized} setting, where the fact is queried directly and the relevant entity is explicitly named (e.g., ``What is the capital of Germany?'').
Such formulations minimize linguistic and discourse complexity, but they abstract away from how facts are accessed in natural language, where entities are frequently introduced indirectly and facts must be retrieved through contextual reference (e.g., ``After arriving in Germany, Alex heads to the country's political capital. Which city does Alex go to?'').
This raises an open question: \emph{does factual knowledge remain equally accessible when it is queried through context?}

In this paper, we evaluate \emph{contextually mediated factual recall} by systematically comparing direct factual queries with contextually mediated counterparts that preserve the same underlying fact while introducing referential mediation (\secref{mediated_query}).
We further examine whether factual recall is influenced by name-specific effects by contrasting synthetic, unseen names with common real names across five languages (\secref{name_investigation}).
Our evaluation spans five languages on three model families, allowing us to analyze how contextual mediation interacts with relation type, model scale, and language.
Overall, our results show that contextual mediation consistently degrades factual recall, with substantial variation across relations, while larger models exhibit greater robustness.
In contrast, the influence of real names and name origin is mixed and unsystematic, suggesting that factual recall under contextual mediation is driven primarily by surrounding context rather than name surface forms.

\section{Related Work}

Recent work shows that LLMs are sensitive to contextual information and can be distracted by irrelevant or misleading content \citep{Petroni2020context,Shi2023irrelevant}.
Such effects are amplified in long contexts due to positional biases and competition for attention \citep{Mohtashami2023infinite,liu-etal-2024-lost,Modarressi2025nolima}, and when contextual information conflicts with a model's parametric knowledge \citep{longpre-etal-2021-entity,chen-etal-2022-rich,wang2023surveyfactualitylargelanguage,xu2024knowledgeconflictsllmssurvey}.

LLMs store and recall factual knowledge \citep{petroni-etal-2019-language}, which has been extensively studied using probing and prompt-based evaluations in English \citep{roberts2020much,peng2022copen} and in multilingual settings \citep{jiang-etal-2020-x,kassner-etal-2021-multilingual,yin-etal-2022-geomlama}.
Several studies show that knowledge acquisition correlates with exposure frequency during pretraining \citep{liu-etal-2025-tracing,Merullo2025frequency}.
Recent work also examines crosslingual consistency and attributes inconsistent recall to representational and entity-level misalignment \citep{qi-etal-2023-cross,wang2025lostmultilingualitydissectingcrosslingual,lu-etal-2025-paths,liu2025entitylevelalignmentcrosslingualconsistency}.

Unlike prior work that evaluates factual recall through direct queries or treats context as optional or distracting, we study \emph{contextually mediated factual recall}, where the target entity is introduced indirectly through context.
This setting allows us to assess factual robustness under naturalistic contextual mediation rather than isolated fact retrieval.

\section{General Experimental Setup}

\textbf{Models} 
We evaluate 9 LLMs of different sizes from 3 model families: \textbf{LLaMA} \citep{grattafiori2024llama3herdmodels}, \textbf{Qwen} \citep{yang2025qwen3technicalreport}, and \textbf{Gemma} \citep{gemmateam2025gemma3technicalreport}.
All models are trained on highly multilingual corpora. 
See details in \secref{env}.

\textbf{Dataset}
We use KLAR \citep{wang2025lostmultilingualitydissectingcrosslingual}, a multilingual factual knowledge probing dataset, in this research.
Our evaluation covers \textbf{1,742} facts from \textbf{9} relation types (cf.\ Table~\ref{tab:relation_fact_counts} in \secref{klar}). 
We use language-specific prompt templates for each relation provided by KLAR for our direct factual knowledge queries.

\textbf{Languages.}
Our evaluation covers five typologically diverse languages spanning multiple writing systems: English (\textbf{EN}), Arabic (\textbf{AR}), Japanese (\textbf{JA}), Korean (\textbf{KO}), and Chinese (\textbf{ZH}).

\textbf{Evaluation.}
We evaluate factual recall using exact-match accuracy.
Models are prompted with three in-context examples followed by the test query, and are allowed to generate up to 10 tokens.
A prediction is considered correct if the gold answer appears as a prefix of the generated output.

\section{Contextually Mediated Factual Recall}\seclabel{mediated_query}

This section introduces our setup for evaluating \emph{contextually mediated factual recall} and contrasts it with standard direct factual queries.

\subsection{Methodology}

In a contextually mediated factual query, factual access is mediated by a short naturalistic context rather than an explicit entity-centric question.
Each query consists of two components: (\textbf{i}) a \emph{context sentence} that embeds the target entity in a realistic scenario, and (\textbf{ii}) a \emph{query sentence} that asks for a factual attribute of that entity.
A placeholder \emph{name} appears across the two sentences and serves as a bridge to maintain discourse coherence.

Formally, let $e$ denote the target entity and $r$ a factual relation.
A contextually mediated query takes the form:
$
c(e, n) \; \Vert \; q(r, n),
$
where $c$ introduces $e$ indirectly through a context involving a name $n$, and $q$ query the object of relation $r$ by referring to the same name.
Correct factual recall requires resolving the referent introduced in the context before recalling the factual relation.

\textbf{Synthetical Names}
To minimize name-based biases, we use \emph{synthetic names} that are unlikely to be associated with real individuals.
Names can encode demographic, cultural, or frequency-based signals that influence model behavior independently of factual knowledge \citep{wei2024uncoveringnamebasedbiaseslarge,manchanda-shivaswamy-2025-name}.
To control for this factor,\footnote{We analyze the effect of real names from different languages separately in \secref{name_investigation}.}
we query a commercial model (i.e., \texttt{ChatGPT-5.2}) to generate candidate synthetic names.
We then filter these candidates using Infini-gram \citep{liu2025infinigramscalingunboundedngram}, which allows querying $n$-gram frequencies across large corpora.\footnote{\url{https://infini-gram.readthedocs.io/}}
Only names that do not occur in any of the indexed corpora are retained, ensuring minimal prior exposure during pretraining.

\begin{figure*}[t]
\setlength{\belowcaptionskip}{-0.3cm}
    \centering
    \begin{subfigure}[t]{0.19\textwidth}
        \centering
        \includegraphics[width=\linewidth]{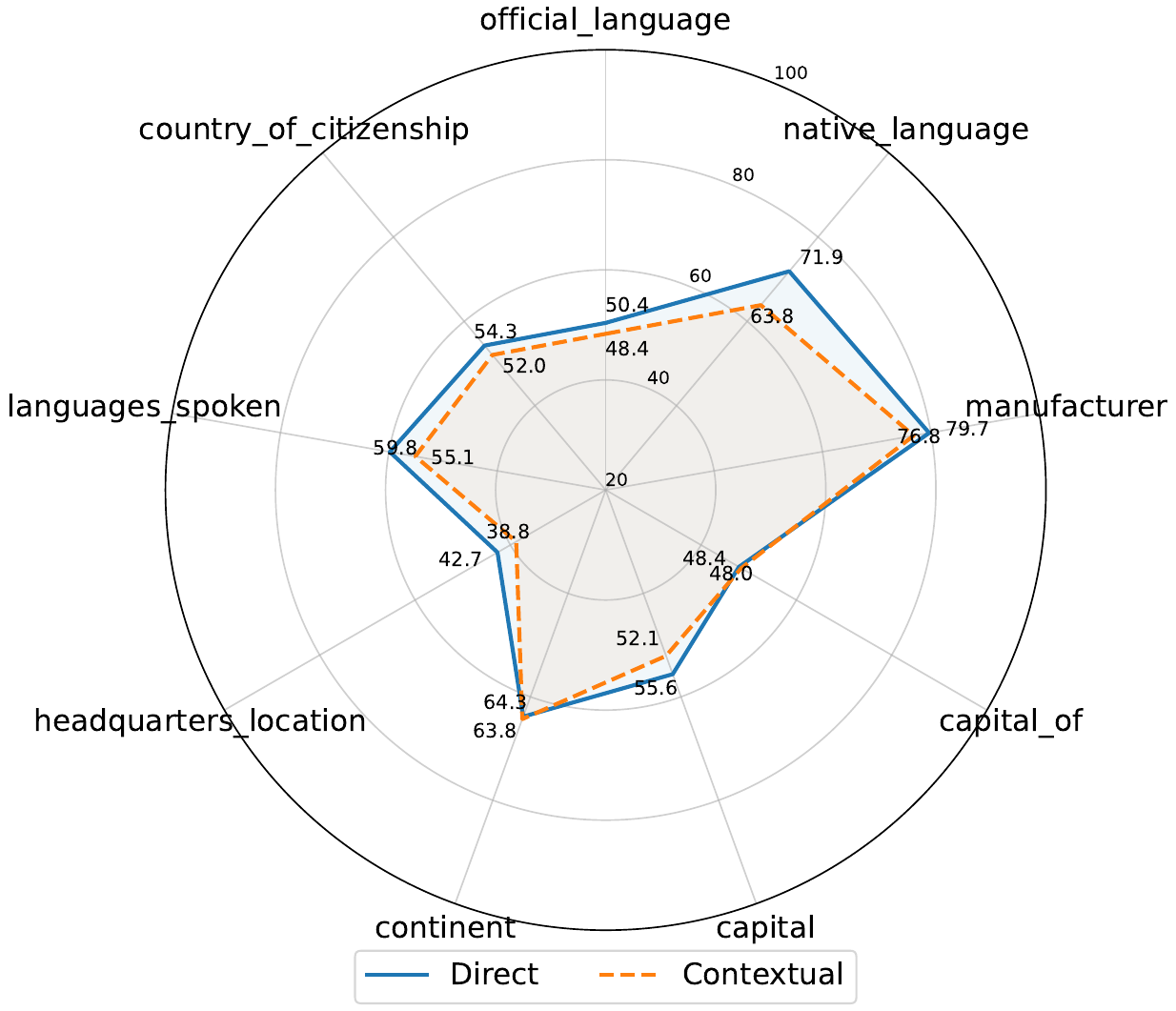}
        \caption{Arabic}
        \label{fig:radar-ar}
    \end{subfigure}
    \hfill
    \begin{subfigure}[t]{0.19\textwidth}
        \centering
        \includegraphics[width=\linewidth]{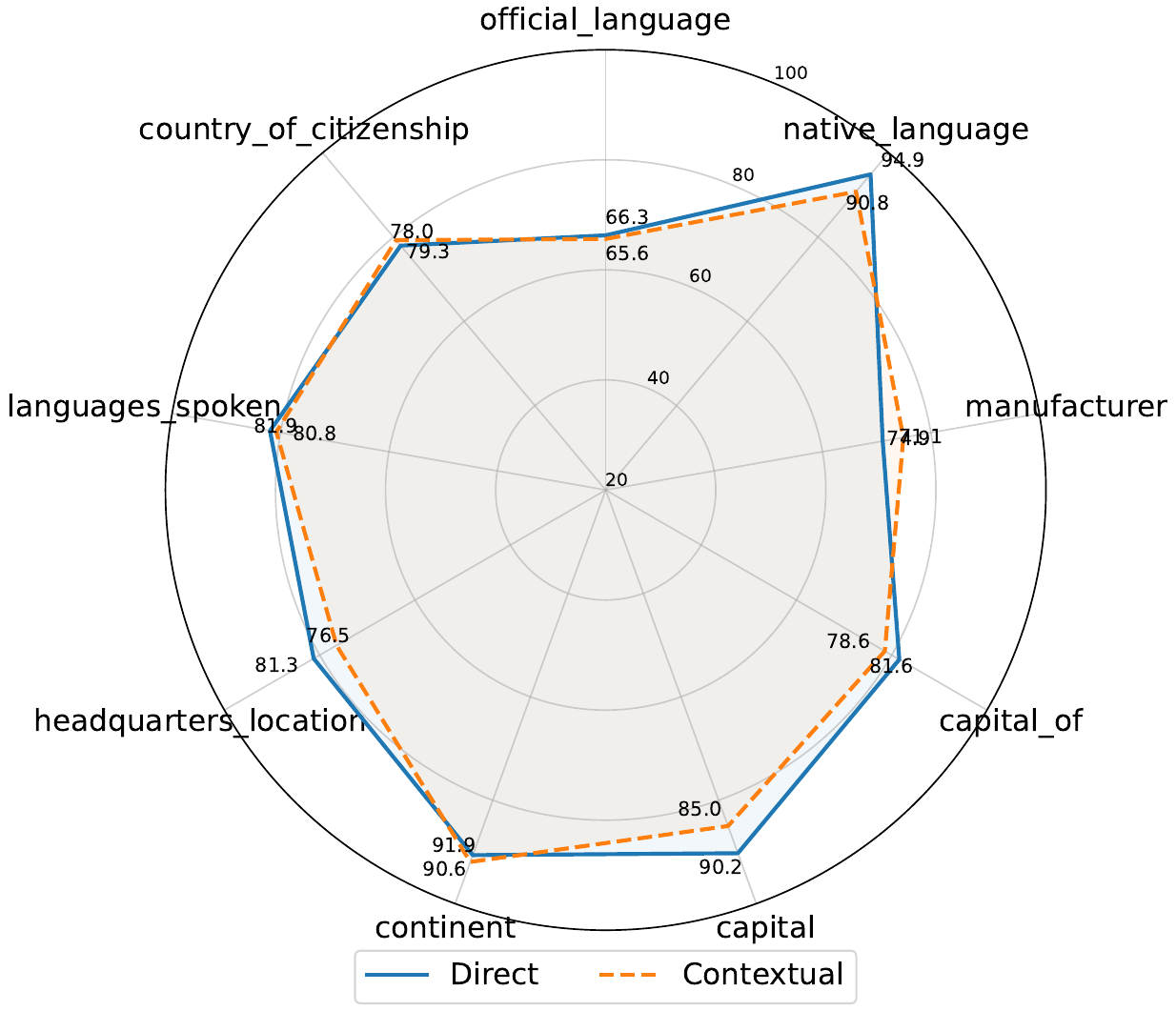}
        \caption{English}
        \label{fig:radar-en}
    \end{subfigure}
    \hfill
    \begin{subfigure}[t]{0.19\textwidth}
        \centering
        \includegraphics[width=\linewidth]{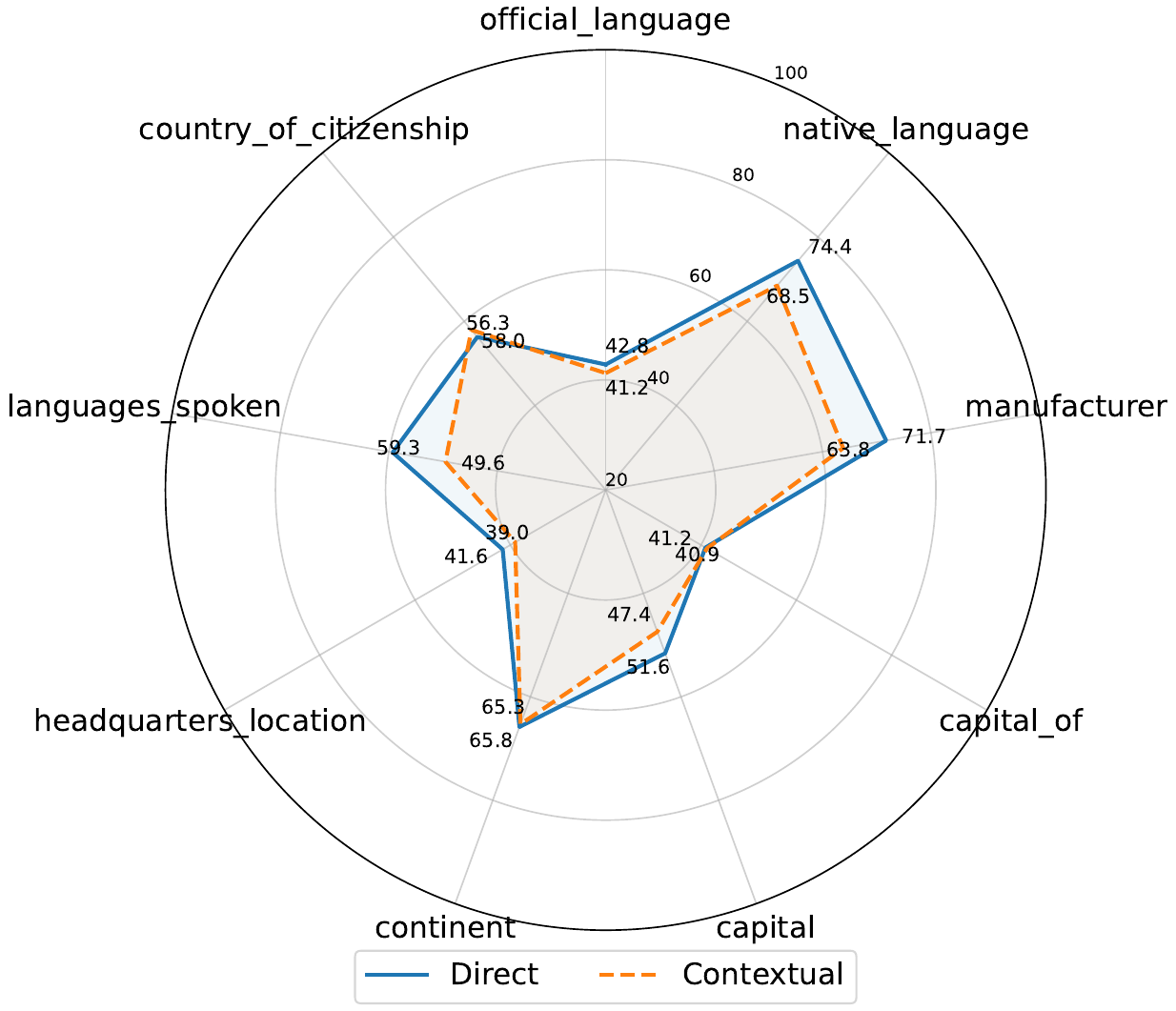}
        \caption{Japanese}
        \label{fig:radar-ja}
    \end{subfigure}
    \hfill
    \begin{subfigure}[t]{0.19\textwidth}
        \centering
        \includegraphics[width=\linewidth]{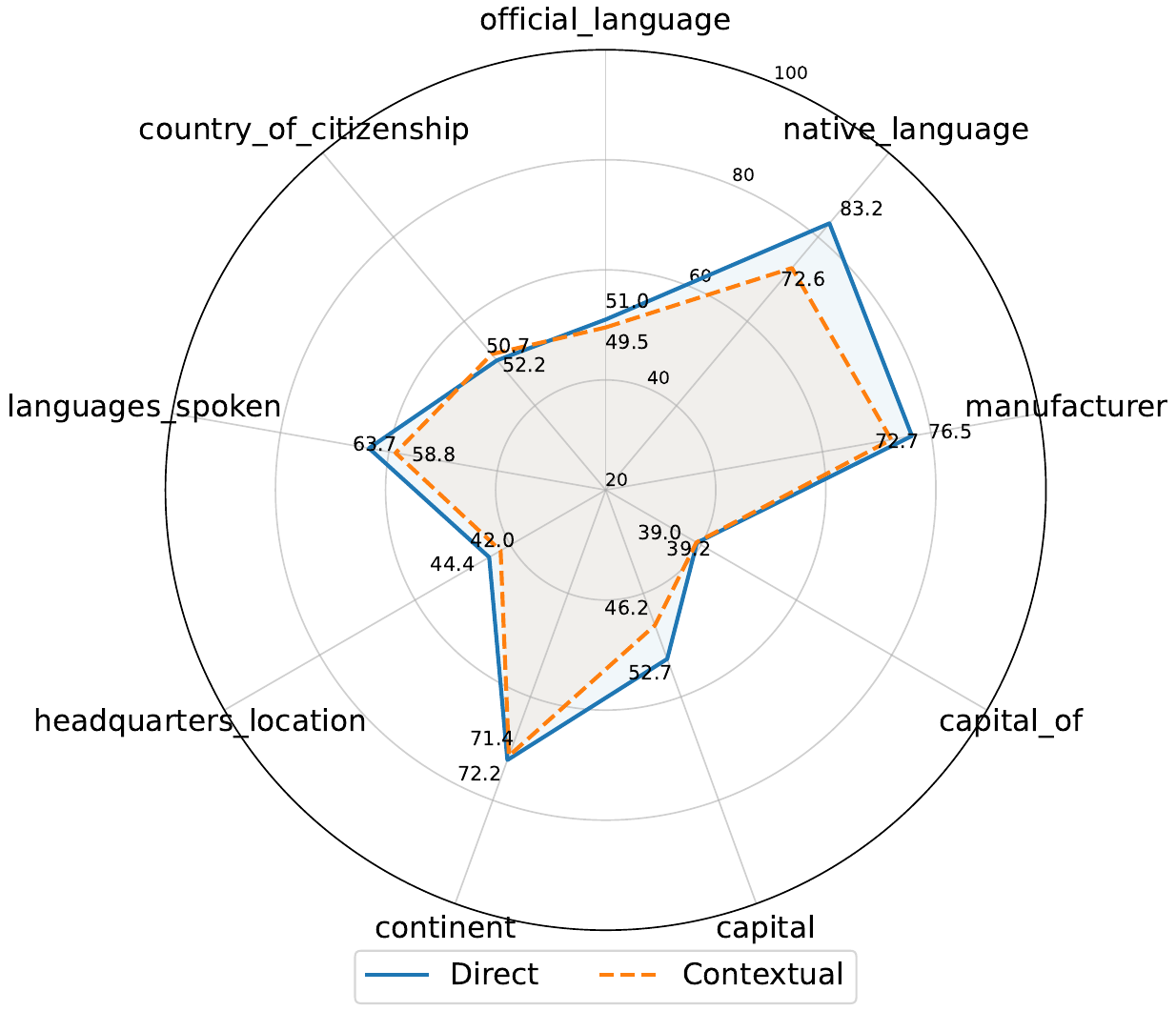}
        \caption{Korean}
        \label{fig:radar-ko}
    \end{subfigure}
    \hfill
    \begin{subfigure}[t]{0.19\textwidth}
        \centering
        \includegraphics[width=\linewidth]{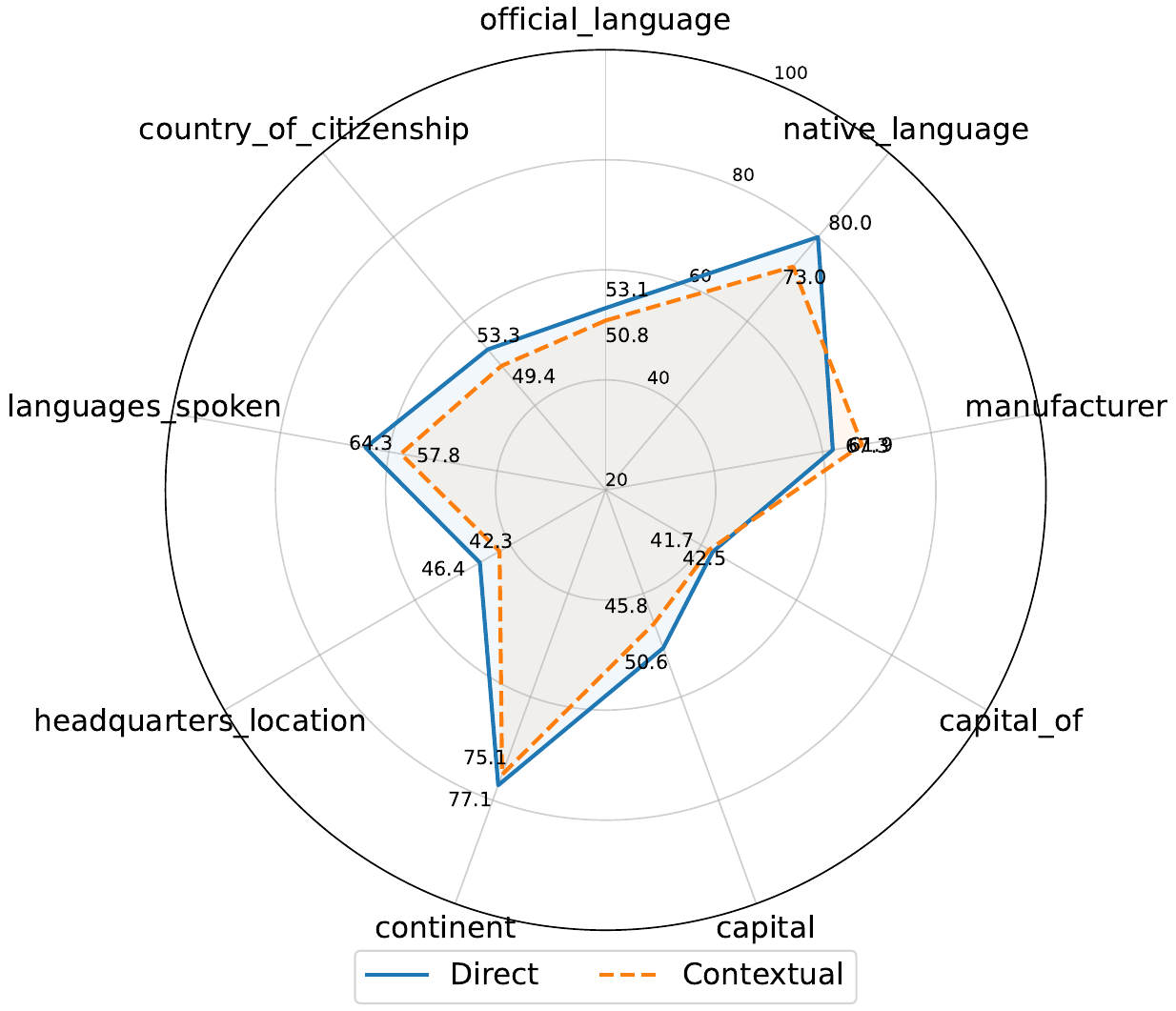}
        \caption{Chinese}
        \label{fig:radar-zh}
    \end{subfigure}
    \caption{
    Per-relation factual recall performance for five languages under \emph{direct factual queries} and \emph{contextually mediated queries}.
    Each radar plot shows accuracy averaged across all models.
    Across languages, introducing contextual mediation generally leads to a relation-dependent performance degradation compared to direct queries.
    }
    \label{fig:radar-all-langs}
\end{figure*}

\begin{figure}[t]
    \setlength{\belowcaptionskip}{-0.4cm}
    \centering
    \includegraphics[width=\linewidth]{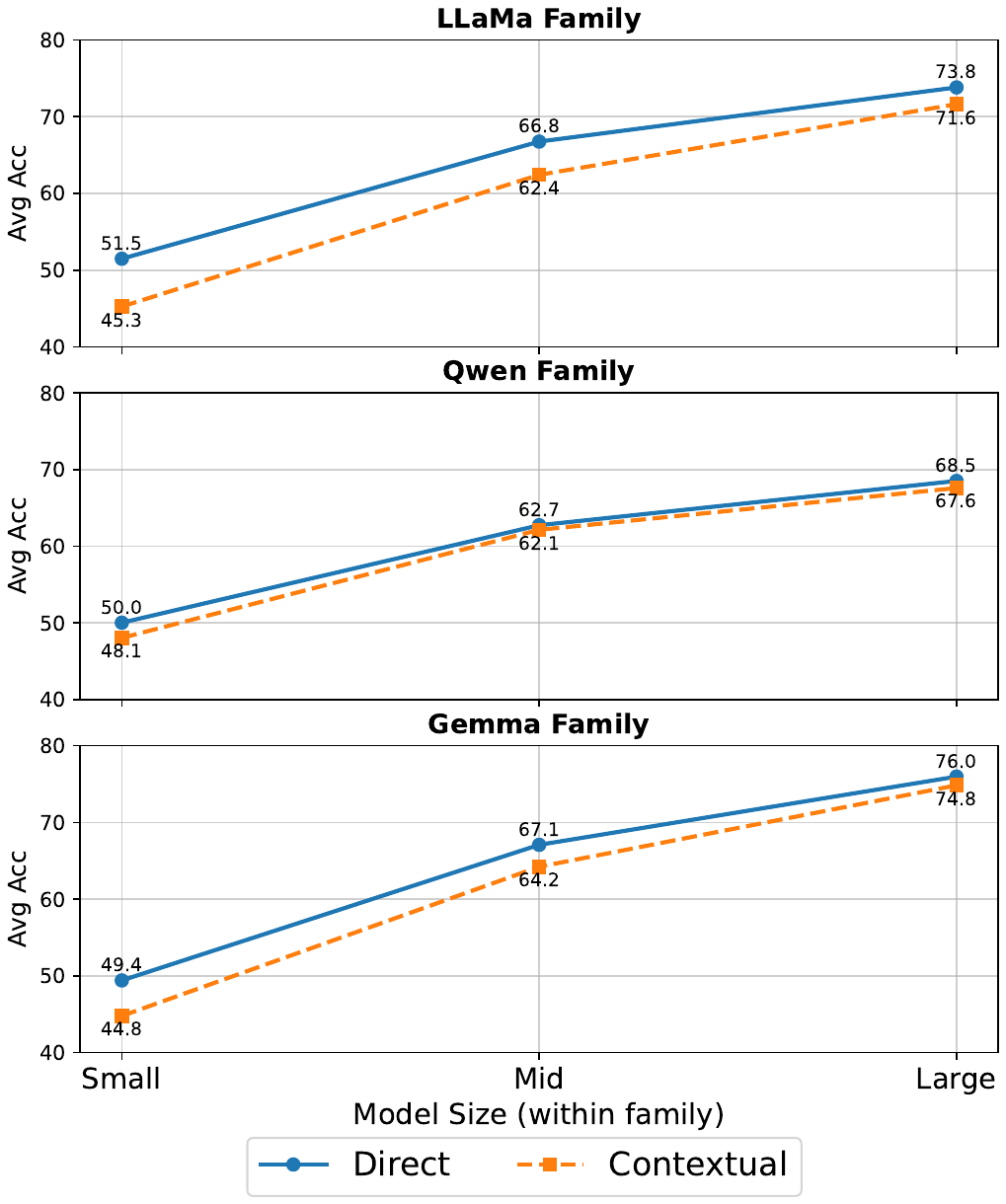}
    \caption{
    Average factual recall performance across relations as a function of model size for each model family.
    Larger models in the LLaMA and Gemma families exhibit increased robustness to contextual mediation, while the Qwen family shows a weaker size effect.
    }
    \label{fig:size}
\end{figure}

\textbf{Prompt Format}
We design two-sentence prompts for each factual relation in English, where the context sentence introduces the target entity implicitly, and the query sentence requests the associated fact.
For example, for the \emph{official language} relation, we use:\footnote{See \secref{templates} for complete templates for all relations.}
\begin{quote}
\emph{``\texttt{\{name\}} is traveling to \texttt{\{subject\}} for a business trip and needs to complete local government forms. Which de jure official language should \texttt{\{name\}} use on the paperwork?''}
\end{quote}
We construct analogous two-sentence templates for other relations, where the underlying fact remains identical to direct factual queries.
The templates are then translated into the other four languages.

\subsection{Results and Discussion}

Figure~\ref{fig:radar-all-langs} reports per-relation factual recall under direct and contextually mediated queries across languages, allowing us to analyze how introducing indirect contextual references affects factual recall.
Figure~\ref{fig:size} further examines whether model scale influences robustness to contextual mediation.

\textbf{Contextual mediation generally reduces factual recall, with strong relation-level variation.}
Across all languages, contextually mediated queries lead to lower factual recall than direct queries.
However, the impact varies substantially by relation.
Relations such as \texttt{native\_language}, \texttt{capital}, and \texttt{headquarters\_location} are consistently the most sensitive, while others, notably \texttt{continent}, remain largely stable.
We hypothesize that this robustness is related to the limited candidate space of certain relations, which reduces ambiguity even under indirect entity access \citep{liu-etal-2025-relation-specific}.
In some cases, contextual mediation can even improve performance (e.g., \texttt{manufacturer} in English and Chinese), suggesting that contextual cues may occasionally aid disambiguation.
We further present an error breakdown in \secref{error}.

\textbf{Larger models exhibit increased robustness to contextual mediation.}
As expected, overall factual recall improves with model scale for both direct and contextually mediated queries.
More importantly, larger models tend to be more robust to contextual mediation, as reflected by a reduced performance gap between direct and mediated settings.
This trend is clearly observed in the LLaMA and Gemma families.
In contrast, the Qwen family shows weaker or less consistent gains in robustness with scale.
These findings suggest that increased model capacity improves not only factual knowledge coverage but also the ability to recover relevant facts when entity access is indirect.
This may reflect stronger internal representations of entities and relations, or more effective integration of contextual cues in larger models.

\section{Influence of Real-Name Bias}\seclabel{name_investigation}

In \secref{mediated_query}, we use synthetic names to minimize potential biases arising from names seen during pretraining.
In this section, we explicitly examine whether replacing synthetic names with \emph{real} names further influences contextually mediated factual recall.

\subsection{Setup}

\begin{figure*}[t]
\setlength{\belowcaptionskip}{-0.3cm}
    \centering
    \begin{subfigure}[t]{0.19\textwidth}
        \centering
        \includegraphics[width=\linewidth]{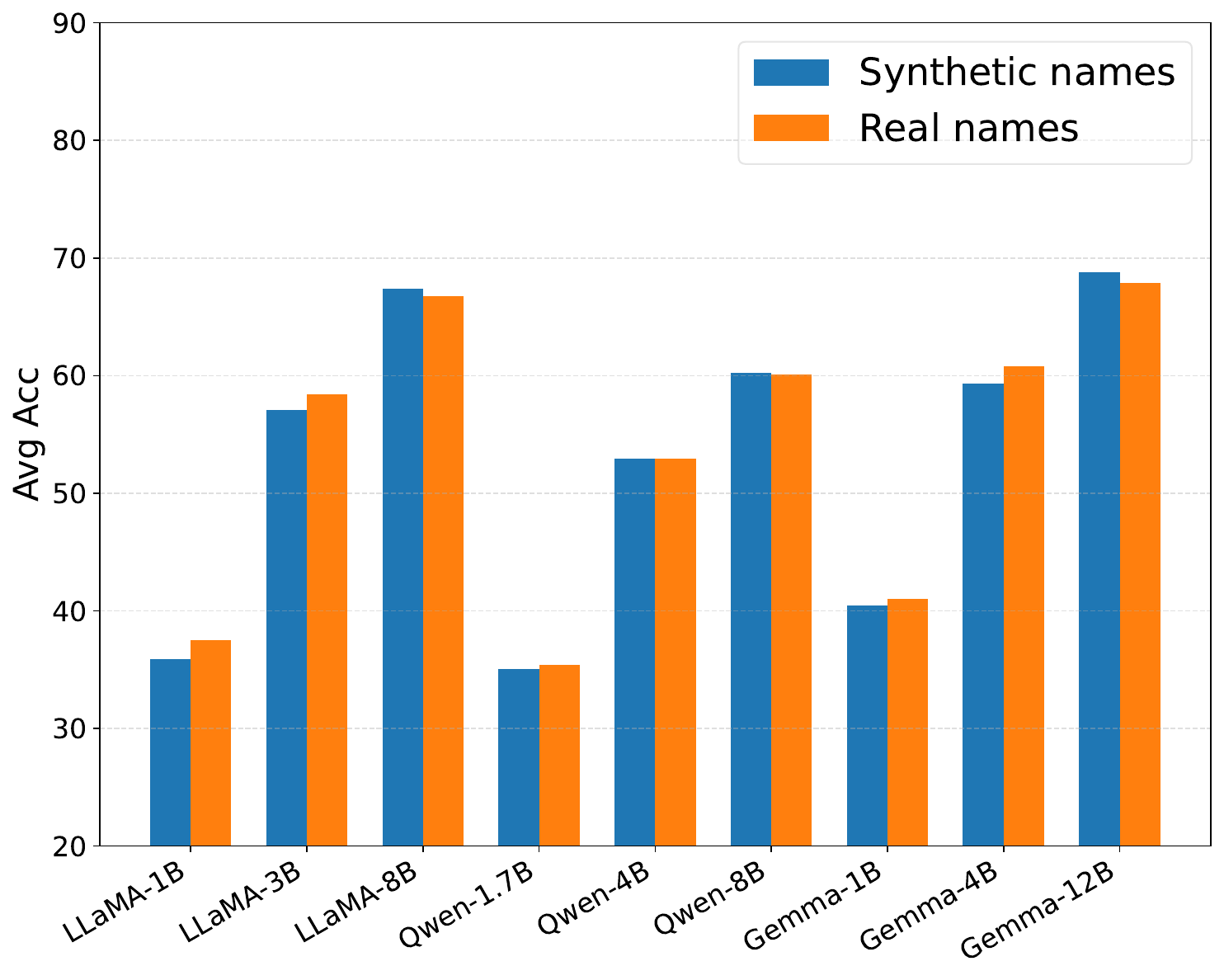}
        \caption{Arabic}
        \label{fig:bar-ar}
    \end{subfigure}
    \hfill
    \begin{subfigure}[t]{0.19\textwidth}
        \centering
        \includegraphics[width=\linewidth]{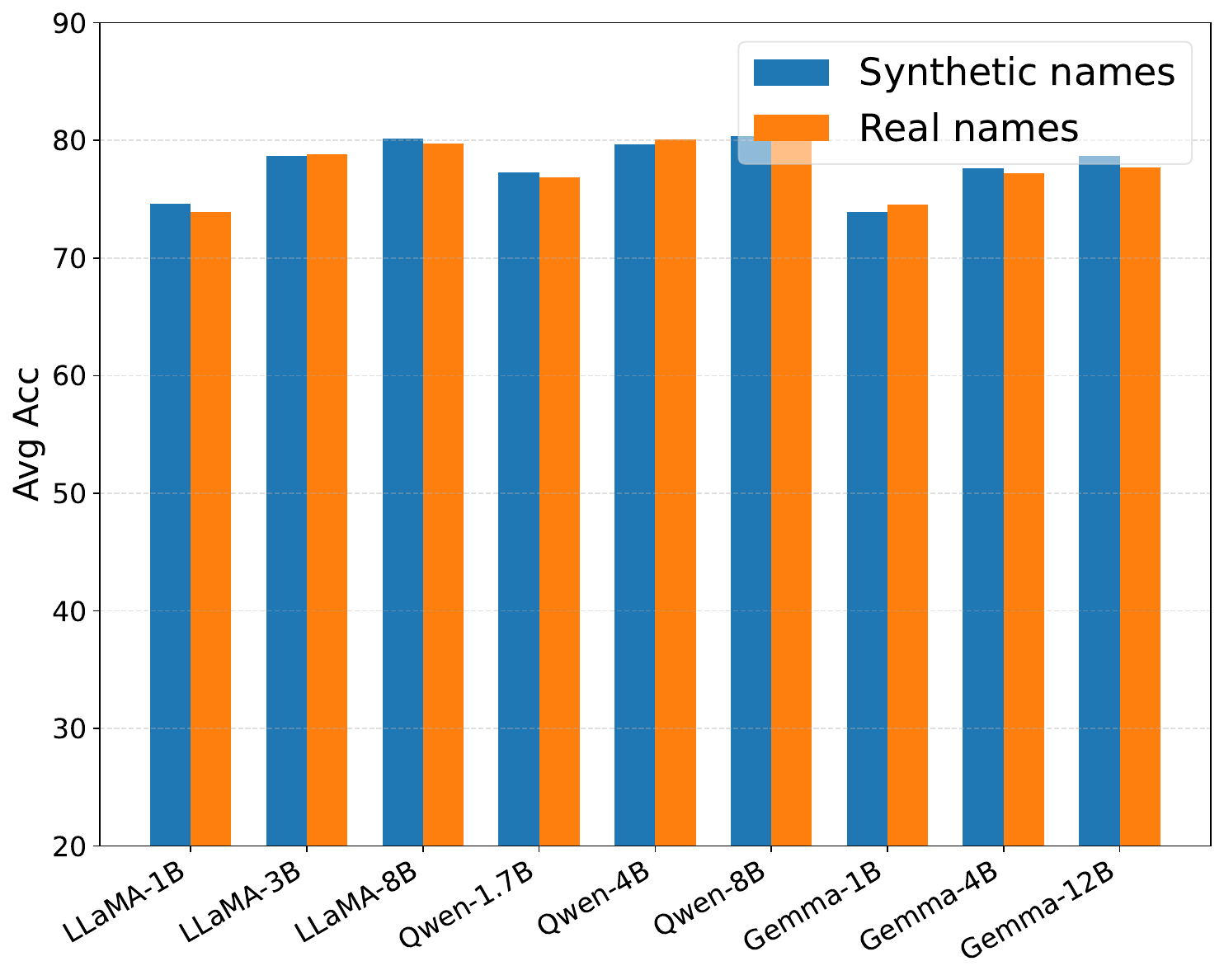}
        \caption{English}
        \label{fig:bar-en}
    \end{subfigure}
    \hfill
    \begin{subfigure}[t]{0.19\textwidth}
        \centering
        \includegraphics[width=\linewidth]{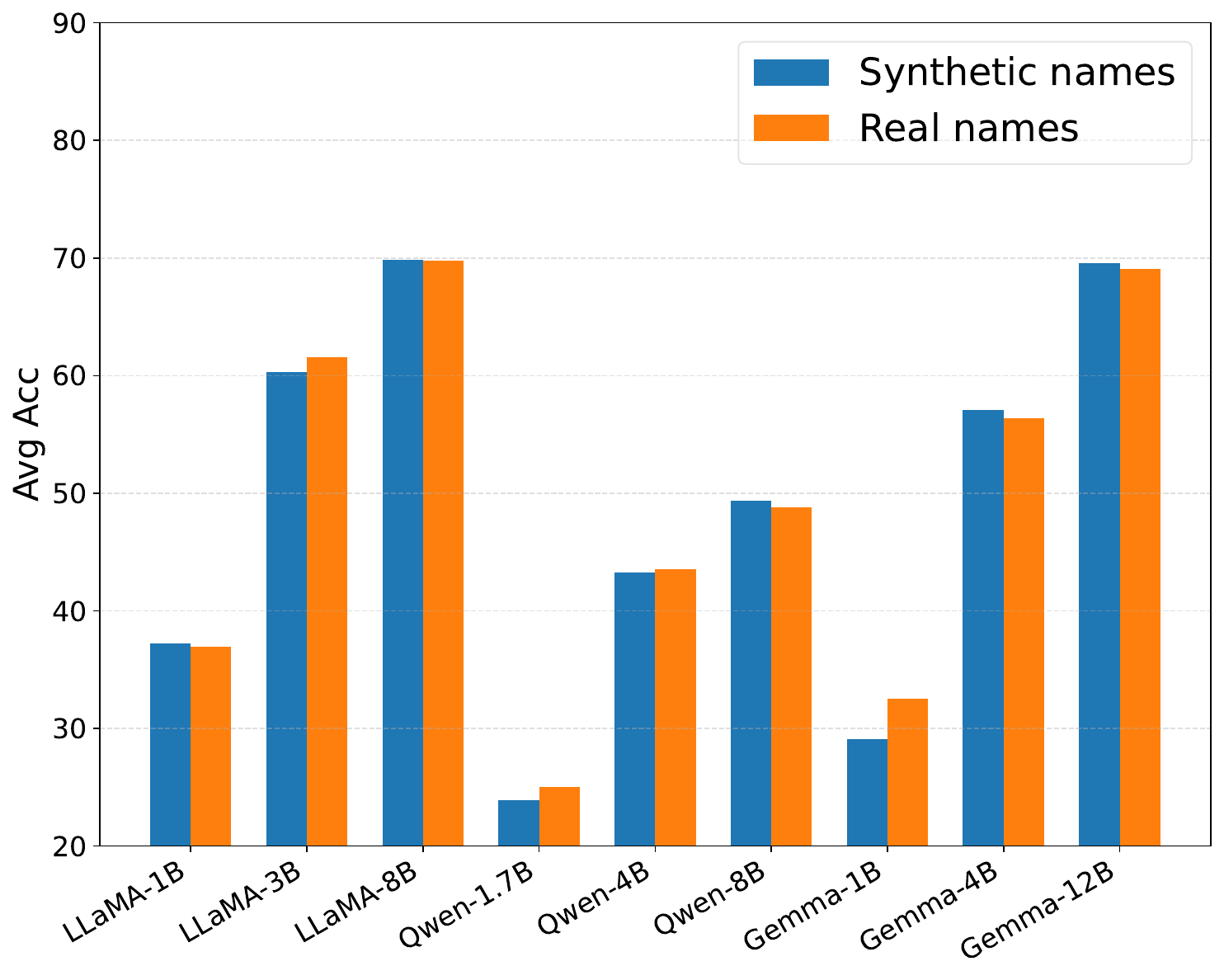}
        \caption{Japanese}
        \label{fig:bar-ja}
    \end{subfigure}
    \hfill
    \begin{subfigure}[t]{0.19\textwidth}
        \centering
        \includegraphics[width=\linewidth]{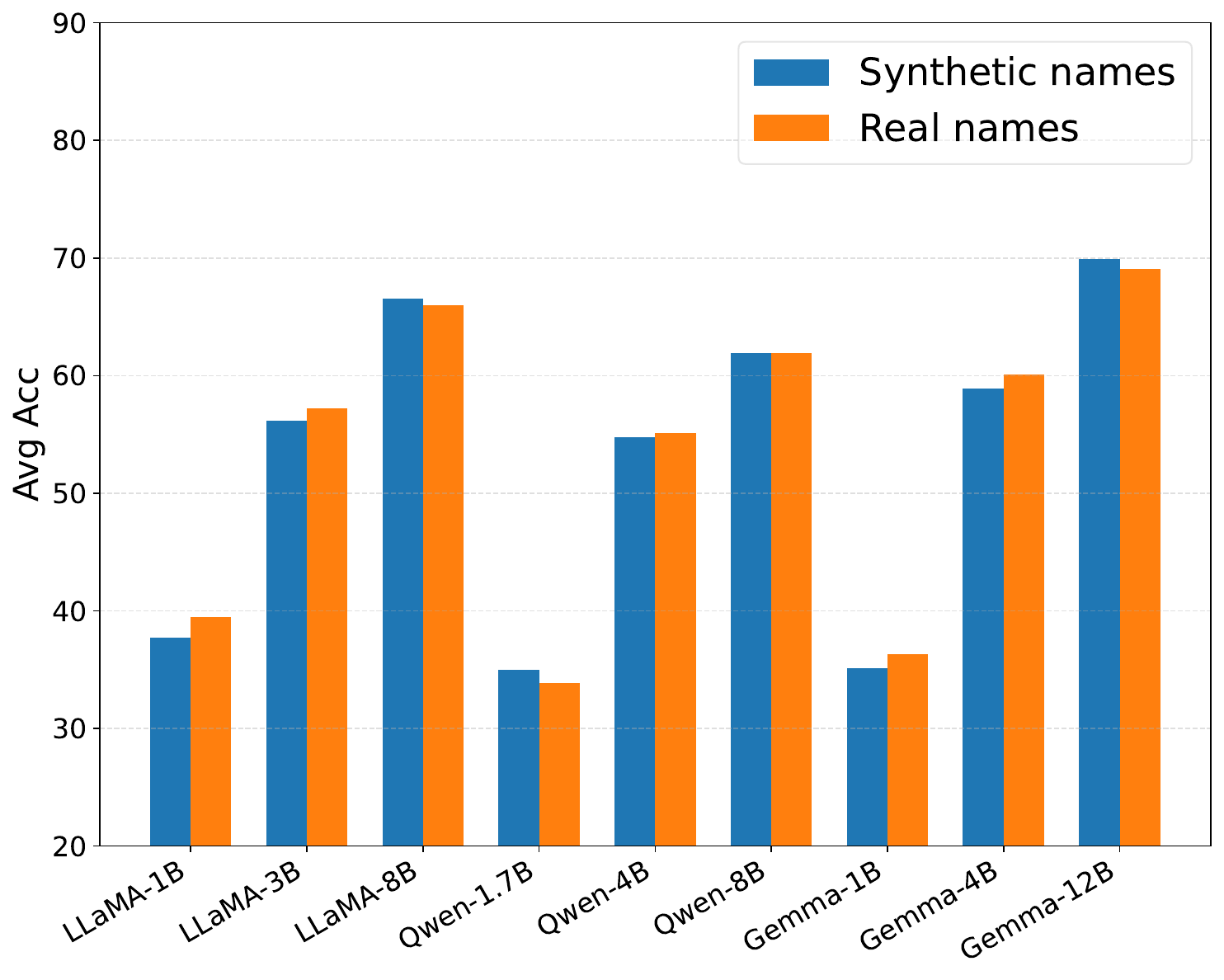}
        \caption{Korean}
        \label{fig:bar-ko}
    \end{subfigure}
    \hfill
    \begin{subfigure}[t]{0.19\textwidth}
        \centering
        \includegraphics[width=\linewidth]{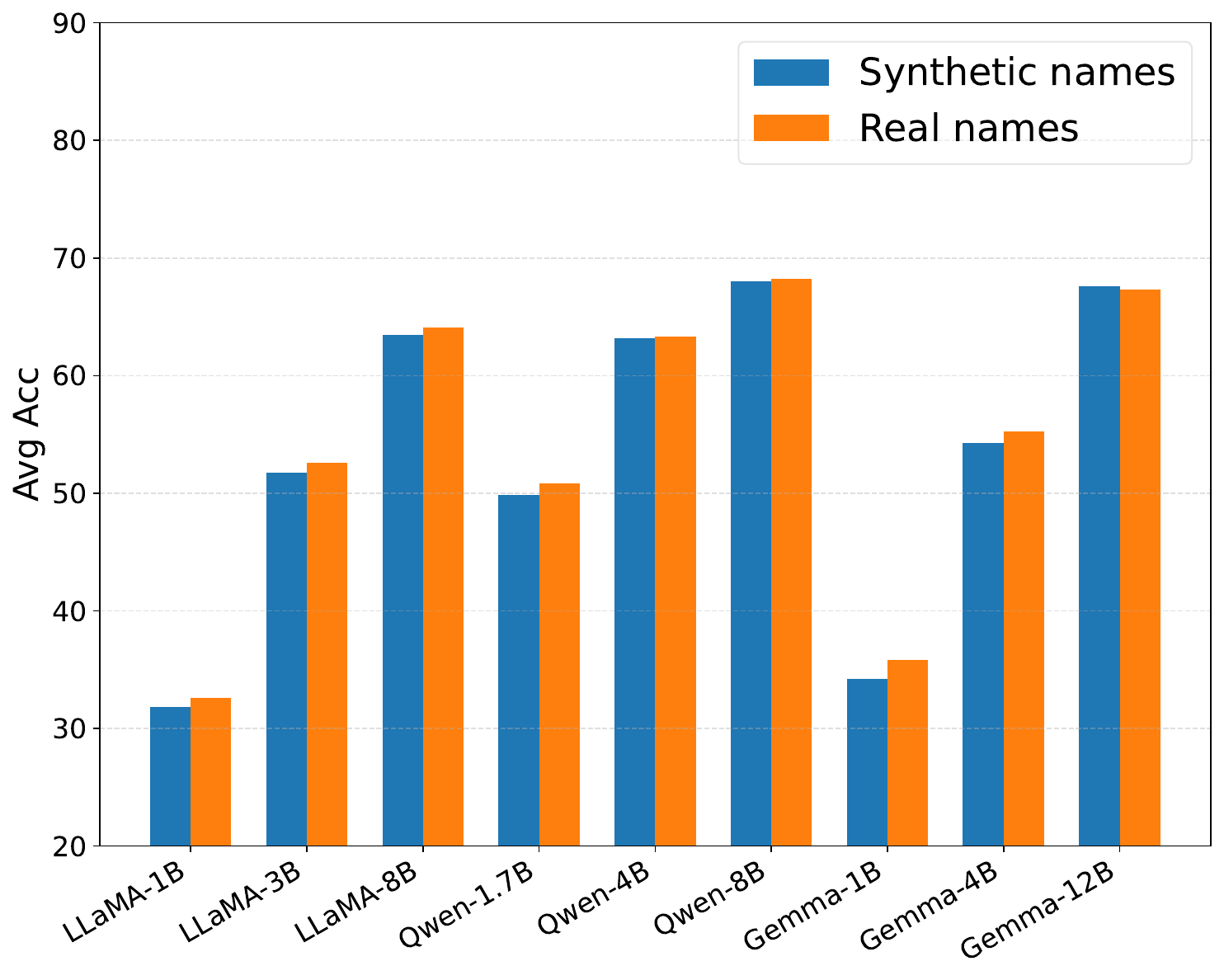}
        \caption{Chinese}
        \label{fig:bar-zh}
    \end{subfigure}
    \caption{
    Per-language comparison of contextually mediated factual recall using synthetic versus real names, averaged across relations for each model.
    The effect of replacing synthetic names with real names is highly mixed across both languages and models, showing no consistent increase or decrease in performance.
    }
    \label{fig:name_bias_per_language}
\end{figure*}

\begin{table}[t]
\setlength{\belowcaptionskip}{-0.3cm}
\renewcommand{\arraystretch}{0.8} 
\centering
\resizebox{\linewidth}{!}{
\begin{tabular}{lccccc}
\toprule
Model & EN & AR & JA & KO & ZH \\
\midrule
LLaMA-1B & 73.9 & 74.1 & 73.8 & 73.4 & 73.8 \\
LLaMA-3B & 78.8 & 78.0 & 78.9 & 78.6 & 78.8 \\
LLaMA-8B & 79.7 & 79.3 & 79.8 & 80.0 & 79.7 \\
\hline
Qwen-1.7B & 76.9 & 76.6 & 77.2 & 76.9 & 76.6 \\
Qwen-4B & 80.1 & 79.0 & 80.1 & 79.7 & 79.9 \\
Qwen-8B & 80.1 & 79.9 & 80.1 & 79.9 & 79.6 \\
\hline
Gemma-3-1B & 74.5 & 73.5 & 74.2 & 74.3 & 74.1 \\
Gemma-3-4B & 77.2 & 77.0 & 77.7 & 77.5 & 77.4 \\
Gemma-3-12B & 77.7 & 77.8 & 78.5 & 78.1 & 77.8 \\
\bottomrule
\end{tabular}
}
\caption{
English contextually mediated factual recall with real names from different language origins.
Each cell reports overall accuracy averaged across relations.
Across model families and sizes, performance remains largely stable across name origins, indicating no systematic degradation or advantage from non-English names.
}
\label{tab:english_queries}
\end{table}

\textbf{Real Names.}
For each considered language, we collect 50 common personal names (25 female and 25 male).
Each name is represented both in its \emph{original} script and in a \emph{Latin transliteration}.

\textbf{Per-Language Queries.}
We construct contextually mediated factual queries using the same templates as in \secref{mediated_query}, but replace synthetic names with real names from the corresponding language, written in their original script.
We compare the resulting performance against the synthetic-name condition obtained in \secref{mediated_query} to assess whether real names introduce additional within-language biases.

\textbf{English Queries with Crosslingual Names.}
To further probe the interaction between name origin and contextual mediation, we insert transliterated real names from non-English languages into the English prompt templates used in \secref{mediated_query}.
This setting allows us to test whether names associated with different cultural backgrounds differentially affect English contextually mediated factual recall, where model performance is typically strongest.

\subsection{Results and Discussion}

Figure~\ref{fig:name_bias_per_language} compares contextually mediated factual recall when using synthetic versus real names across languages.
Table~\ref{tab:english_queries} further reports English contextually mediated factual recall using transliterated real names from five language origins.

\textbf{Real names do not introduce a systematic performance trend.}
Across models and languages, replacing synthetic names with real names does not lead to a consistent increase or decrease in factual recall.
While real names occasionally yield small performance gains (e.g., for Chinese names in the LLaMA family), these effects are neither stable across models nor consistent across languages.
Overall, the influence of real names appears highly heterogeneous,
However, there is no systematic trend,
suggesting that name-related biases interact with model-specific and language-specific factors rather than introducing a uniform effect.

\textbf{Name origin plays a limited role compared to contextual mediation.}
Using transliterated real names from different language origins in English prompts results in only marginal performance variation.
For example, across LLaMA, Qwen, and Gemma models in Table~\ref{tab:english_queries}, accuracies remain within a narrow range (typically within $\approx$1\%) when replacing English names with Arabic, Japanese, Korean, or Chinese names.
Together with the per-language analysis, this suggests that under contextual mediation, factual recall is driven primarily by the \emph{surrounding context} rather than by the specific surface form or linguistic origin of the name.

\section{Conclusion}

We study \emph{contextually mediated factual recall}, a setting where factual knowledge must be retrieved through indirect contextual cues rather than explicit entity mentions.
Across five languages and multiple model families, we find that introducing contextual mediation generally degrades factual recall performance, with substantial variation across relations.
We further show that larger models are more resilient to contextual mediation.
Finally, we find no systematic evidence that real names introduce consistent bias under contextual mediation, indicating that surrounding context plays a more dominant role than name surface forms.
Overall, our results highlight important limitations of current factual recall evaluations and motivate the need for more context-aware and multilingual benchmarks.

\section*{Limitations}

While this study provides insights into contextually mediated factual recall, it has several limitations.

First, we focus on a fixed set of relations and prompt templates, which may not fully capture the diversity and complexity of contextual references encountered in natural language use.

Second, our analysis of name bias is based on a limited set of common names and does not consider sociolinguistic factors such as name frequency or variation across demographic groups.

Finally, although we evaluate multiple open-source model families across different scales, our analysis excludes closed-source models, which may exhibit different or more robust behavior under contextual mediation.

\bibliography{custom}

\appendix

\newpage

\section{KLAR Dataset Details}\seclabel{klar}

The statistics of the used subset of the KLAR dataset \citep{wang2025lostmultilingualitydissectingcrosslingual} are presented in Table~\ref{tab:relation_fact_counts}.
KLAR is based on BMLAMA17 \citep{qi-etal-2023-cross} that covers 17 languages and contains 20 relation types. 
We use \textbf{1,742} facts grouped into \textbf{9}~relation categories and \textbf{5} typologically different languages.

\begin{table}[t]
\small
\setlength{\belowcaptionskip}{-0.4cm}
\setlength{\tabcolsep}{0.1mm}
\centering
\resizebox{\linewidth}{!}{%
\begin{tabular}{l r}
\hline
\textbf{Relation} & \textbf{Number of Facts} \\
\hline
\texttt{capital} & 336 \\
\texttt{capital\_of} & 212 \\
\texttt{continent} & 212 \\
\texttt{country\_of\_citizenship} & 60 \\
\texttt{headquarters\_location} & 51 \\
\texttt{languages\_spoken} & 104 \\
\texttt{manufacturer} & 35 \\
\texttt{native\_language} & 130 \\
\texttt{official\_language} & 602 \\
\hline
\textbf{Total} & \textbf{1,742} \\
\hline
\end{tabular}
}
\caption{Number of facts per relation type.}
\label{tab:relation_fact_counts}
\end{table}

\section{Complete Results}

We present complete per-relation factual recall results covering both direct and contextually mediated queries, for all evaluated models and languages in Table~\ref{tab:relperf:en:v0}, \ref{tab:relperf:ar:v0}, \ref{tab:relperf:ja:v0}, \ref{tab:relperf:ko:v0}, and \ref{tab:relperf:zh:v0}.

\begin{table}[t]
\setlength{\belowcaptionskip}{-0.3cm}
\centering
\resizebox{\linewidth}{!}{
\begin{tabular}{lccc}
\toprule
Model & \texttt{native\_language} & \texttt{capital} & \texttt{headquarters\_location} \\
\midrule
LLaMA-1B  & 22 & 33 & 10 \\
LLaMA-3B  & 4  & 14 & 4  \\
LLaMA-8B  & 6  & 13 & 0  \\
\hline
Qwen-1.7B & 7  & 23 & 3  \\
Qwen-4B   & 3  & 10 & 5  \\
Qwen-8B   & 2  & 26 & 2  \\
\hline
Gemma-3-1B  & 12 & 30 & 7  \\
Gemma-3-4B  & 6  & 45 & 4  \\
Gemma-3-12B & 2  & 43 & 2  \\
\bottomrule
\end{tabular}
}
\caption{
Counts of \emph{contextual-mediation failures} on English for the most susceptible relations.
Each entry reports the number of facts where a model answers the direct query correctly but fails on the corresponding contextually mediated query.
}
\label{tab:mediation_failures_en}
\end{table}

\section{Error Breakdown}\seclabel{error}

We conduct an error analysis to manually check the relations that are most susceptible to contextual mediation in \secref{mediated_query}, namely \texttt{native\_language}, \texttt{capital}, and \texttt{headquarters\_location}. 
Specifically, we focus on facts for which the model correctly answers the direct query but fails under the corresponding contextually mediated formulation, allowing us to isolate errors attributable to contextual mediation rather than missing factual knowledge.

We report the number of such facts for each model and relation in Table~\ref{tab:mediation_failures_en}.
Overall, the number of contextual-mediation failures decreases with model scale, most clearly for \texttt{native\_language} and \texttt{headquarters\_location}, reinforcing our earlier finding that larger models are generally more robust to contextual mediation (cf.~\secref{mediated_query}).
In contrast, the effect of scale on \texttt{capital} is less consistent: while LLaMA models exhibit a clear reduction in failures as size increases, both Qwen and Gemma show non-monotonic trends, with larger models even incurring more errors than their smaller counterparts.
This suggests that robustness to contextual mediation is relation-dependent and cannot be explained by model scale alone, pointing to differences in how factual associations are represented and accessed across model families.

\section{Prompt Templates}\seclabel{templates}

We present the prompt template for direct factual recall and contextually mediated factual recall we used in our study for English in Table~\ref{tab:prompt}.
The prompt templates used for the other four languages (i.e., Arabic, Japanese, Korean, Chinese) are translations of the English prompt templates, which we create using \texttt{ChatGPT-5.2} and then have native speakers manually verify.

\begin{table*}[t]
\centering
\small
\setlength{\tabcolsep}{6pt}
\begin{tabular}{llp{10.5cm}}
\toprule
Relation & Query Type & Prompt Template \\
\midrule

\multirow{2}{*}{\textbf{Official language}}
& Direct &
What is the official language of \texttt{\{subject\}}? \\
& Contextual &
\texttt{\{name\}} is traveling to \texttt{\{subject\}} for a business trip and needs to complete local government forms.
Which de jure official language should \texttt{\{name\}} use on the paperwork? \\

\midrule
\multirow{2}{*}{\textbf{Native language}}
& Direct &
What is the native language of \texttt{\{subject\}}? \\
& Contextual &
\texttt{\{name\}} finds a private letter \texttt{\{subject\}} wrote to their family as a child.
Assuming it was written in the author's native language, what language does \texttt{\{name\}} see the letter is written in? \\

\midrule
\multirow{2}{*}{\textbf{Manufacturer}}
& Direct &
Which company manufactures \texttt{\{subject\}}? \\
& Contextual &
\texttt{\{name\}} picks up \texttt{\{subject\}} to contact customer service.
Which company should \texttt{\{name\}} contact as the manufacturer? \\

\midrule
\multirow{2}{*}{\textbf{Capital of}}
& Direct &
What is the country or state for which \texttt{\{subject\}} is the capital? \\
& Contextual &
\texttt{\{name\}} arrives in the city of \texttt{\{subject\}}.
Which country or state is this city the capital of? \\

\midrule
\multirow{2}{*}{\textbf{Capital}}
& Direct &
Where is \texttt{\{subject\}}'s capital located? \\
& Contextual &
After arriving in \texttt{\{subject\}}, \texttt{\{name\}} heads to the country's political capital.
Which city does \texttt{\{name\}} go to? \\

\midrule
\multirow{2}{*}{\textbf{Continent}}
& Direct &
Which continent is \texttt{\{subject\}} located in? \\
& Contextual &
\texttt{\{name\}} is planning a trip to \texttt{\{subject\}}.
Which continent will \texttt{\{name\}} be traveling to? \\

\midrule
\multirow{2}{*}{\textbf{Headquarters location}}
& Direct &
Where is the headquarters of \texttt{\{subject\}}? \\
& Contextual &
\texttt{\{name\}} walks into the headquarters lobby of \texttt{\{subject\}}.
Which city is \texttt{\{name\}} in right now? \\

\midrule
\multirow{2}{*}{\textbf{Languages spoken}}
& Direct &
What language did \texttt{\{subject\}} use to communicate? \\
& Contextual &
\texttt{\{name\}} receives a voice message from \texttt{\{subject\}}.
Which language does \texttt{\{name\}} hear in the message? \\

\midrule
\multirow{2}{*}{\textbf{Country of citizenship}}
& Direct &
Which country is \texttt{\{subject\}} a citizen of? \\
& Contextual &
\texttt{\{name\}} is at border control checking \texttt{\{subject\}}'s documents.
Which country of citizenship does \texttt{\{name\}} see listed? \\

\bottomrule
\end{tabular}
\caption{Prompt templates for direct and contextually mediated factual recall across relations in \textbf{English}.
\texttt{\{subject\}} denotes the target entity and \texttt{\{name\}} denotes a (synthetic or real) person name used for contextual mediation.}
\label{tab:prompt}
\end{table*}

\section{Models, Hyperparameters, and Environment}\seclabel{env}

\subsection{Models}

We evaluate a collection of 9 decoder-only LLMs drawn from four prominent model families: \textbf{LLaMA} \citep{grattafiori2024llama3herdmodels}, \textbf{Qwen} \citep{yang2025qwen3technicalreport}, and \textbf{Gemma} \citep{gemmateam2025gemma3technicalreport}.
LLaMA, Qwen, and Gemma are trained on broadly multilingual corpora.
Within the LLaMA family, we evaluate \texttt{Llama-3.2-1B}, \texttt{Llama-3.2-3B}, and \texttt{Llama-3.1-8B}.
For Qwen, we include \texttt{Qwen3-1.7B-Base}, \texttt{Qwen3-4B-Base}, and \texttt{Qwen3-8B-Base}.
The Gemma models considered are \texttt{gemma-3-1b-pt}, \texttt{gemma-3-4b-pt}, and \texttt{gemma-3-12b-pt}.

\subsection{Hyperparameters}

All experiments use greedy decoding with a 3-shot prompting setup, where in-context examples are randomly sampled from the same relation.
We fix the random seed to 12345 for reproducibility.

\subsection{Environment}

All experiments are executed using the vLLM framework.\footnote{\url{https://vllm.ai/}}
Experiments are run on NVIDIA RTX A6000 GPUs with 48\,GB of memory.

\begin{table*}[t]
\setlength{\belowcaptionskip}{-0.3cm}
\renewcommand{\arraystretch}{0.85}
\centering
\resizebox{\linewidth}{!}{%
\begin{tabular}{llccccccccc}
\toprule
Relation & Query & LLaMA-1B & LLaMA-3B & LLaMA-8B & Qwen-1.7B & Qwen-4B & Qwen-8B & Gemma-3-1B & Gemma-3-4B & Gemma-3-12B \\
\midrule
\multirow{2}{*}{\texttt{official\_language}} & Direct & 66.3 & 67.9 & 67.8 & 65.6 & 64.6 & 65.0 & 65.3 & 67.8 & 66.3 \\
 & Contextual & 64.1 & 65.1 & 66.3 & 65.3 & 65.0 & 68.4 & 62.0 & 67.8 & 66.6 \\
\hline
\addlinespace[2pt]
\multirow{2}{*}{\texttt{native\_language}} & Direct & 93.8 & 96.2 & 95.4 & 93.8 & 95.4 & 95.4 & 90.0 & 95.4 & 98.5 \\
 & Contextual & 79.2 & 93.8 & 92.3 & 90.0 & 94.6 & 95.4 & 82.3 & 92.3 & 96.9 \\
\hline
\addlinespace[2pt]
\multirow{2}{*}{\texttt{manufacturer}} & Direct & 74.3 & 74.3 & 71.4 & 65.7 & 74.3 & 71.4 & 65.7 & 71.4 & 71.4 \\
 & Contextual & 74.3 & 77.1 & 77.1 & 74.3 & 77.1 & 71.4 & 71.4 & 77.1 & 74.3 \\
\hline
\addlinespace[2pt]
\multirow{2}{*}{\texttt{capital\_of}} & Direct & 76.4 & 81.6 & 83.0 & 78.3 & 84.4 & 84.0 & 79.2 & 82.5 & 84.9 \\
 & Contextual & 77.4 & 76.9 & 78.3 & 75.0 & 83.5 & 81.6 & 77.4 & 77.4 & 79.7 \\
\hline
\addlinespace[2pt]
\multirow{2}{*}{\texttt{capital}} & Direct & 90.2 & 90.8 & 92.0 & 88.1 & 89.6 & 90.5 & 87.5 & 91.1 & 92.6 \\
 & Contextual & 83.0 & 88.7 & 89.9 & 85.7 & 90.5 & 85.4 & 81.8 & 79.5 & 80.4 \\
\hline
\addlinespace[2pt]
\multirow{2}{*}{\texttt{continent}} & Direct & 86.8 & 91.5 & 93.4 & 89.6 & 89.2 & 93.4 & 90.6 & 89.6 & 91.5 \\
 & Contextual & 91.5 & 91.5 & 93.4 & 92.0 & 91.5 & 93.9 & 89.6 & 90.1 & 93.9 \\
\hline
\addlinespace[2pt]
\multirow{2}{*}{\texttt{headquarters\_location}} & Direct & 76.5 & 90.2 & 90.2 & 70.6 & 72.5 & 86.3 & 66.7 & 86.3 & 92.2 \\
 & Contextual & 64.7 & 84.3 & 90.2 & 70.6 & 66.7 & 84.3 & 56.9 & 80.4 & 90.2 \\
\hline
\addlinespace[2pt]
\multirow{2}{*}{\texttt{languages\_spoken}} & Direct & 80.8 & 83.7 & 81.7 & 83.7 & 84.6 & 78.8 & 80.8 & 82.7 & 80.8 \\
 & Contextual & 70.2 & 81.7 & 83.7 & 80.8 & 84.6 & 82.7 & 78.8 & 82.7 & 81.7 \\
\hline
\addlinespace[2pt]
\multirow{2}{*}{\texttt{country\_of\_citizenship}} & Direct & 71.7 & 80.0 & 81.7 & 73.3 & 76.7 & 78.3 & 80.0 & 81.7 & 78.3 \\
 & Contextual & 70.0 & 78.3 & 85.0 & 80.0 & 83.3 & 85.0 & 70.0 & 80.0 & 81.7 \\
\hline
\bottomrule
\end{tabular}}
\caption{Per-relation accuracy for direct and contextually mediated queries. Language: \textbf{English}. }
\label{tab:relperf:en:v0}
\end{table*}

\begin{table*}[t]
\setlength{\belowcaptionskip}{-0.3cm}
\renewcommand{\arraystretch}{0.85}
\centering
\resizebox{\linewidth}{!}{%
\begin{tabular}{llccccccccc}
\toprule
Relation & Query & LLaMA-1B & LLaMA-3B & LLaMA-8B & Qwen-1.7B & Qwen-4B & Qwen-8B & Gemma-3-1B & Gemma-3-4B & Gemma-3-12B \\
\midrule
\multirow{2}{*}{\texttt{official\_language}} & Direct & 41.4 & 55.8 & 60.1 & 37.2 & 48.5 & 55.0 & 37.9 & 55.3 & 62.1 \\
 & Contextual & 31.7 & 51.2 & 60.1 & 33.6 & 49.5 & 55.8 & 36.2 & 55.3 & 62.0 \\
\hline
\addlinespace[2pt]
\multirow{2}{*}{\texttt{native\_language}} & Direct & 60.0 & 76.9 & 90.0 & 49.2 & 66.9 & 79.2 & 55.4 & 76.9 & 92.3 \\
 & Contextual & 35.4 & 73.1 & 77.7 & 36.9 & 73.1 & 76.9 & 43.8 & 71.5 & 86.2 \\
\hline
\addlinespace[2pt]
\multirow{2}{*}{\texttt{manufacturer}} & Direct & 74.3 & 82.9 & 85.7 & 68.6 & 80.0 & 82.9 & 65.7 & 88.6 & 88.6 \\
 & Contextual & 57.1 & 71.4 & 82.9 & 68.6 & 82.9 & 88.6 & 62.9 & 85.7 & 91.4 \\
\hline
\addlinespace[2pt]
\multirow{2}{*}{\texttt{capital\_of}} & Direct & 29.7 & 52.8 & 59.4 & 23.1 & 47.6 & 55.7 & 35.4 & 58.5 & 69.3 \\
 & Contextual & 27.8 & 51.4 & 64.2 & 24.1 & 46.7 & 58.0 & 34.0 & 58.0 & 71.7 \\
\hline
\addlinespace[2pt]
\multirow{2}{*}{\texttt{capital}} & Direct & 47.9 & 64.6 & 73.8 & 29.8 & 48.2 & 58.6 & 45.2 & 61.0 & 71.4 \\
 & Contextual & 45.5 & 61.6 & 70.5 & 27.1 & 48.5 & 55.1 & 39.6 & 56.5 & 64.3 \\
\hline
\addlinespace[2pt]
\multirow{2}{*}{\texttt{continent}} & Direct & 49.1 & 67.9 & 76.4 & 50.5 & 64.6 & 71.2 & 50.0 & 69.8 & 75.0 \\
 & Contextual & 43.4 & 69.3 & 77.4 & 54.7 & 63.7 & 70.3 & 56.6 & 67.9 & 75.5 \\
\hline
\addlinespace[2pt]
\multirow{2}{*}{\texttt{headquarters\_location}} & Direct & 35.3 & 41.2 & 52.9 & 25.5 & 37.3 & 45.1 & 37.3 & 45.1 & 64.7 \\
 & Contextual & 17.6 & 43.1 & 58.8 & 21.6 & 35.3 & 43.1 & 17.6 & 49.0 & 62.7 \\
\hline
\addlinespace[2pt]
\multirow{2}{*}{\texttt{languages\_spoken}} & Direct & 42.3 & 63.5 & 73.1 & 43.3 & 61.5 & 70.2 & 43.3 & 65.4 & 76.0 \\
 & Contextual & 33.7 & 52.9 & 72.1 & 42.3 & 51.9 & 62.5 & 45.2 & 60.6 & 75.0 \\
\hline
\addlinespace[2pt]
\multirow{2}{*}{\texttt{country\_of\_citizenship}} & Direct & 40.0 & 50.0 & 70.0 & 41.7 & 51.7 & 68.3 & 45.0 & 51.7 & 70.0 \\
 & Contextual & 33.3 & 43.3 & 66.7 & 40.0 & 51.7 & 63.3 & 45.0 & 53.3 & 71.7 \\
\hline
\bottomrule
\end{tabular}}
\caption{Per-relation accuracy for direct and contextually mediated queries. Language: \textbf{Arabic}. }
\label{tab:relperf:ar:v0}
\end{table*}

\begin{table*}[t]
\setlength{\belowcaptionskip}{-0.3cm}
\renewcommand{\arraystretch}{0.85}
\centering
\resizebox{\linewidth}{!}{%
\begin{tabular}{llccccccccc}
\toprule
Relation & Query & LLaMA-1B & LLaMA-3B & LLaMA-8B & Qwen-1.7B & Qwen-4B & Qwen-8B & Gemma-3-1B & Gemma-3-4B & Gemma-3-12B \\
\midrule
\multirow{2}{*}{\texttt{official\_language}} & Direct & 35.5 & 56.5 & 60.1 & 15.8 & 35.9 & 38.5 & 29.9 & 51.3 & 61.6 \\
 & Contextual & 29.1 & 53.8 & 62.5 & 15.8 & 35.4 & 38.5 & 27.7 & 50.0 & 58.3 \\
\hline
\addlinespace[2pt]
\multirow{2}{*}{\texttt{native\_language}} & Direct & 69.2 & 83.1 & 90.8 & 52.3 & 74.6 & 67.7 & 60.8 & 78.5 & 92.3 \\
 & Contextual & 52.3 & 78.5 & 86.9 & 42.3 & 70.0 & 68.5 & 50.8 & 76.2 & 90.8 \\
\hline
\addlinespace[2pt]
\multirow{2}{*}{\texttt{manufacturer}} & Direct & 60.0 & 82.9 & 80.0 & 57.1 & 68.6 & 74.3 & 54.3 & 82.9 & 85.7 \\
 & Contextual & 51.4 & 65.7 & 71.4 & 48.6 & 60.0 & 68.6 & 51.4 & 80.0 & 77.1 \\
\hline
\addlinespace[2pt]
\multirow{2}{*}{\texttt{capital\_of}} & Direct & 25.0 & 49.1 & 62.7 & 10.8 & 36.8 & 50.0 & 16.5 & 49.1 & 68.4 \\
 & Contextual & 22.6 & 50.5 & 65.1 & 10.4 & 34.4 & 50.5 & 15.6 & 53.3 & 68.9 \\
\hline
\addlinespace[2pt]
\multirow{2}{*}{\texttt{capital}} & Direct & 45.8 & 68.5 & 76.5 & 17.6 & 40.8 & 52.7 & 27.4 & 61.0 & 74.1 \\
 & Contextual & 39.3 & 64.3 & 76.2 & 14.0 & 41.1 & 47.3 & 22.3 & 54.2 & 68.2 \\
\hline
\addlinespace[2pt]
\multirow{2}{*}{\texttt{continent}} & Direct & 51.9 & 79.7 & 83.5 & 57.1 & 59.4 & 66.0 & 35.4 & 74.5 & 84.9 \\
 & Contextual & 62.3 & 76.9 & 78.3 & 58.0 & 56.1 & 62.7 & 37.7 & 70.3 & 85.4 \\
\hline
\addlinespace[2pt]
\multirow{2}{*}{\texttt{headquarters\_location}} & Direct & 31.4 & 51.0 & 60.8 & 21.6 & 37.3 & 45.1 & 21.6 & 43.1 & 62.7 \\
 & Contextual & 21.6 & 39.2 & 54.9 & 25.5 & 35.3 & 45.1 & 19.6 & 45.1 & 64.7 \\
\hline
\addlinespace[2pt]
\multirow{2}{*}{\texttt{languages\_spoken}} & Direct & 51.9 & 71.2 & 81.7 & 31.7 & 45.2 & 50.0 & 51.9 & 75.0 & 75.0 \\
 & Contextual & 37.5 & 53.8 & 71.2 & 19.2 & 43.3 & 48.1 & 34.6 & 61.5 & 76.9 \\
\hline
\addlinespace[2pt]
\multirow{2}{*}{\texttt{country\_of\_citizenship}} & Direct & 48.3 & 61.7 & 63.3 & 38.3 & 53.3 & 68.3 & 48.3 & 53.3 & 71.7 \\
 & Contextual & 43.3 & 65.0 & 68.3 & 40.0 & 60.0 & 71.7 & 36.7 & 58.3 & 78.3 \\
\hline
\bottomrule
\end{tabular}}
\caption{Per-relation accuracy for direct and contextually mediated queries. Language: \textbf{Japanese}. }
\label{tab:relperf:ja:v0}
\end{table*}

\begin{table*}[t]
\setlength{\belowcaptionskip}{-0.3cm}
\renewcommand{\arraystretch}{0.85}
\centering
\resizebox{\linewidth}{!}{%
\begin{tabular}{llccccccccc}
\toprule
Relation & Query & LLaMA-1B & LLaMA-3B & LLaMA-8B & Qwen-1.7B & Qwen-4B & Qwen-8B & Gemma-3-1B & Gemma-3-4B & Gemma-3-12B \\
\midrule
\multirow{2}{*}{\texttt{official\_language}} & Direct & 44.0 & 58.0 & 59.3 & 36.0 & 49.8 & 57.8 & 36.4 & 56.0 & 61.5 \\
 & Contextual & 38.7 & 52.7 & 61.3 & 31.1 & 54.3 & 58.3 & 33.1 & 54.2 & 62.3 \\
\hline
\addlinespace[2pt]
\multirow{2}{*}{\texttt{native\_language}} & Direct & 69.2 & 84.6 & 93.8 & 73.1 & 85.4 & 88.5 & 74.6 & 87.7 & 92.3 \\
 & Contextual & 51.5 & 78.5 & 80.0 & 66.9 & 79.2 & 80.8 & 53.1 & 76.9 & 86.9 \\
\hline
\addlinespace[2pt]
\multirow{2}{*}{\texttt{manufacturer}} & Direct & 65.7 & 82.9 & 88.6 & 57.1 & 74.3 & 85.7 & 65.7 & 82.9 & 85.7 \\
 & Contextual & 62.9 & 74.3 & 85.7 & 51.4 & 71.4 & 82.9 & 60.0 & 82.9 & 82.9 \\
\hline
\addlinespace[2pt]
\multirow{2}{*}{\texttt{capital\_of}} & Direct & 17.9 & 44.8 & 50.9 & 16.5 & 37.3 & 49.5 & 17.5 & 48.6 & 69.3 \\
 & Contextual & 18.9 & 41.0 & 53.8 & 17.5 & 38.7 & 48.6 & 17.0 & 47.6 & 68.4 \\
\hline
\addlinespace[2pt]
\multirow{2}{*}{\texttt{capital}} & Direct & 41.7 & 57.7 & 69.6 & 26.2 & 43.5 & 59.8 & 36.6 & 63.7 & 75.3 \\
 & Contextual & 31.8 & 55.1 & 68.2 & 22.0 & 39.6 & 53.0 & 28.6 & 53.3 & 64.0 \\
\hline
\addlinespace[2pt]
\multirow{2}{*}{\texttt{continent}} & Direct & 59.4 & 72.6 & 84.4 & 60.4 & 74.5 & 78.8 & 54.2 & 77.8 & 87.7 \\
 & Contextual & 53.8 & 71.2 & 81.6 & 61.8 & 75.5 & 84.0 & 52.8 & 75.5 & 86.8 \\
\hline
\addlinespace[2pt]
\multirow{2}{*}{\texttt{headquarters\_location}} & Direct & 23.5 & 47.1 & 62.7 & 21.6 & 41.2 & 54.9 & 29.4 & 51.0 & 68.6 \\
 & Contextual & 21.6 & 33.3 & 52.9 & 27.5 & 43.1 & 54.9 & 23.5 & 47.1 & 74.5 \\
\hline
\addlinespace[2pt]
\multirow{2}{*}{\texttt{languages\_spoken}} & Direct & 39.4 & 64.4 & 76.9 & 50.0 & 68.3 & 75.0 & 46.2 & 73.1 & 79.8 \\
 & Contextual & 42.3 & 58.7 & 71.2 & 40.4 & 63.5 & 68.3 & 43.3 & 68.3 & 73.1 \\
\hline
\addlinespace[2pt]
\multirow{2}{*}{\texttt{country\_of\_citizenship}} & Direct & 36.7 & 51.7 & 63.3 & 36.7 & 58.3 & 53.3 & 33.3 & 55.0 & 68.3 \\
 & Contextual & 31.7 & 53.3 & 65.0 & 31.7 & 60.0 & 60.0 & 36.7 & 60.0 & 71.7 \\
\hline
\bottomrule
\end{tabular}}
\caption{Per-relation accuracy for direct and contextually mediated queries. Language: \textbf{Korean}. }
\label{tab:relperf:ko:v0}
\end{table*}

\begin{table*}[t]
\setlength{\belowcaptionskip}{-0.3cm}
\renewcommand{\arraystretch}{0.85}
\centering
\resizebox{\linewidth}{!}{%
\begin{tabular}{llccccccccc}
\toprule
Relation & Query & LLaMA-1B & LLaMA-3B & LLaMA-8B & Qwen-1.7B & Qwen-4B & Qwen-8B & Gemma-3-1B & Gemma-3-4B & Gemma-3-12B \\
\midrule
\multirow{2}{*}{\texttt{official\_language}} & Direct & 35.4 & 53.2 & 61.5 & 50.7 & 59.5 & 63.0 & 39.5 & 54.5 & 60.5 \\
 & Contextual & 27.2 & 45.8 & 58.1 & 50.2 & 59.6 & 65.0 & 36.2 & 53.5 & 61.5 \\
\hline
\addlinespace[2pt]
\multirow{2}{*}{\texttt{native\_language}} & Direct & 64.6 & 78.5 & 88.5 & 83.8 & 92.3 & 90.8 & 52.3 & 80.0 & 89.2 \\
 & Contextual & 47.7 & 70.0 & 80.0 & 80.0 & 90.0 & 88.5 & 48.5 & 65.4 & 86.9 \\
\hline
\addlinespace[2pt]
\multirow{2}{*}{\texttt{manufacturer}} & Direct & 45.7 & 68.6 & 68.6 & 54.3 & 62.9 & 68.6 & 60.0 & 62.9 & 65.7 \\
 & Contextual & 62.9 & 71.4 & 74.3 & 62.9 & 65.7 & 77.1 & 57.1 & 60.0 & 74.3 \\
\hline
\addlinespace[2pt]
\multirow{2}{*}{\texttt{capital\_of}} & Direct & 21.7 & 41.0 & 53.8 & 32.5 & 50.9 & 55.2 & 22.6 & 43.9 & 60.4 \\
 & Contextual & 19.8 & 41.5 & 54.7 & 28.3 & 50.5 & 54.7 & 17.0 & 46.2 & 62.7 \\
\hline
\addlinespace[2pt]
\multirow{2}{*}{\texttt{capital}} & Direct & 29.8 & 51.2 & 67.6 & 34.5 & 51.8 & 65.5 & 34.5 & 55.4 & 64.9 \\
 & Contextual & 23.2 & 48.8 & 63.4 & 31.5 & 47.6 & 58.6 & 29.2 & 47.0 & 62.8 \\
\hline
\addlinespace[2pt]
\multirow{2}{*}{\texttt{continent}} & Direct & 64.6 & 77.8 & 86.3 & 79.7 & 88.2 & 90.6 & 49.1 & 74.5 & 83.0 \\
 & Contextual & 59.4 & 73.1 & 80.2 & 76.4 & 87.7 & 92.0 & 45.3 & 73.1 & 88.2 \\
\hline
\addlinespace[2pt]
\multirow{2}{*}{\texttt{headquarters\_location}} & Direct & 23.5 & 41.2 & 54.9 & 45.1 & 49.0 & 58.8 & 29.4 & 52.9 & 62.7 \\
 & Contextual & 21.6 & 39.2 & 51.0 & 37.3 & 47.1 & 51.0 & 25.5 & 47.1 & 60.8 \\
\hline
\addlinespace[2pt]
\multirow{2}{*}{\texttt{languages\_spoken}} & Direct & 43.3 & 60.6 & 72.1 & 65.4 & 83.7 & 76.0 & 43.3 & 59.6 & 75.0 \\
 & Contextual & 26.9 & 51.9 & 64.4 & 61.5 & 80.8 & 75.0 & 35.6 & 54.8 & 69.2 \\
\hline
\addlinespace[2pt]
\multirow{2}{*}{\texttt{country\_of\_citizenship}} & Direct & 35.0 & 53.3 & 61.7 & 48.3 & 63.3 & 70.0 & 31.7 & 50.0 & 66.7 \\
 & Contextual & 35.0 & 46.7 & 56.7 & 48.3 & 66.7 & 66.7 & 25.0 & 43.3 & 56.7 \\
\hline
\bottomrule
\end{tabular}}
\caption{Per-relation accuracy for direct and contextually mediated queries. Language: \textbf{Chinese}. }
\label{tab:relperf:zh:v0}
\end{table*}

\end{document}